\documentclass[letterpaper, 10 pt]{article}
\usepackage[T1]{fontenc}
\usepackage[utf8]{inputenc}
\usepackage{amsmath}
\usepackage{amsthm}
\usepackage{amsfonts}
\usepackage{amssymb}
\usepackage{graphicx}
\usepackage{epsfig}
\usepackage{epstopdf}
\usepackage{color}
\usepackage{soul}
\usepackage{mathtools}
\usepackage{appendix}
\usepackage{comment}
\usepackage{bbm}
\usepackage[margin=1in]{geometry}
\usepackage[colorlinks, linkcolor=blue, anchorcolor=blue, citecolor=blue]{hyperref}
\usepackage{enumerate}
\usepackage{empheq}
\usepackage[numbers]{natbib}
\usepackage[ruled,vlined]{algorithm2e}
\usepackage{algorithmic}
\usepackage{dsfont}
\allowdisplaybreaks

\title{\LARGE \bf Impact of Data Processing on Fairness in Supervised Learning}

\author{
  Sajad Khodadadian\\
  Georgia Institute of Technology\\
  \texttt{skhodadadian3@gatech.edu}
  \and
  AmirEmad Ghassami\\
  Johns Hopkins University\\
  \texttt{aghassa1@jhu.edu}
  \and
  Negar Kiyavash\\
  {\'E}cole Polytechnique F{\'e}d{\'e}rale de Lausanne (EPFL)\\
  \texttt{negar.kiyavash@epfl.ch}
}

\newtheorem{lem}{Lemma}

\newtheorem{prop}{Proposition}

\theoremstyle{definition}

\newtheorem{defn}{Definition}

\newtheorem{rem}{Remark}

\renewcommand{\tilde}{\widetilde}

\DeclareMathOperator*{\argmin}{\arg\!\min}

\newcommand{\ExpVal}[2]{\mathbb{E}\left[ #2 \right]}

\newcommand{\sto}{\mbox{\normalfont s.t.}}

\newcommand{\EE}[1]{\ExpVal{}{#1}}

\newcommand{\KL}{\textnormal{D}_{\scalebox{.6}{\textnormal KL}}}

\newcommand{\fdiv}{\textnormal{D}_f}
\newcommand{\TV}{\textnormal{D}_{\scalebox{.6}{\textnormal TV}}}
\newcommand{\W}{W_{Y|X}}
\date{\today}
\begin{document}
\maketitle

\begin{abstract}
We study the impact of pre and post processing for reducing discrimination in data-driven decision makers. We first analyze the fundamental trade-off between fairness and accuracy in a pre-processing approach, and propose a design for a pre-processing module based on a convex optimization program, which can be added before the original classifier. This leads to a fundamental lower bound on attainable discrimination, given any acceptable distortion in the outcome. Furthermore, we reformulate an existing post-processing method in terms of our accuracy and fairness measures, which allows comparing post-processing and pre-processing approaches. We show that under some mild conditions, pre-processing outperforms post-processing. Finally, we show that by appropriate choice of the discrimination measure, the optimization problem for both pre and post processing approaches will reduce to a linear program and hence can be solved efficiently.
\end{abstract}

\section{Introduction}

Despite the success of machine learning algorithms in prediction tasks, a number of recent reports have documented the fact that these algorithms may be biased and discriminate against some demographics. These biases affect a wide range of applications such as healthcare \cite{kinyanjui2019estimating}, facial recognition \cite{buolamwini2018gender}, and loan default risk prediction \cite{tan2017detecting}. 

The issue of discrimination can be formalized as follows. Consider a classification task (e.g., predicting whether or not a prisoner will commit a crime after being released from prison), in which the goal is to assign a label to each individual based on a set of features (e.g., age, charge degree). To prevent discrimination, it is desired to exclude \emph{sensitive attributes}, such as race, gender, religion, etc., from influencing the decision maker\footnote{Discrimination with respect to such attributes is prohibited by law. Specifically, the Title VII of the Civil Rights Act of 1964 prohibits employers from discriminating against employees on the basis of such features.}. Discrimination can be caused either directly by feeding the sensitive attribute as an input to the classifier (also known as \textit{disparate treatment}\cite{mendez1979presumptions}), or indirectly, where a sensitive attribute is omitted from the input, but it still affects the prediction through proxy variables, such as education level, geographic location, etc. Indirect discrimination is referred to as \textit{disparate impact} in the literature \cite{barocas2016big}.

The issue of disparate impact has motivated a large body of research on (a) how to identify disparate impact \cite{feldman2015certifying, wang2020split}, (b) how to measure disparate impact \cite{hardt2016equality,berk2018fairness}, (c) how to reduce disparate impact \cite{zafar2017parity,corbett2017algorithmic, agarwal2018reductions}. The approaches for reducing disparate impact, can be categorized as pre-processing \cite{kamiran2012decision, zhang2018mitigating}, in-processing \cite{calders2010three, celis2019classification}, and post-processing\cite{pedreschi2009measuring,hardt2016equality, wei2019optimized}, which correspond to controlled distortion of the training set, modification of the  learning algorithm, and processing the output of the classifier after it has been trained, respectively\cite{romei2014multidisciplinary}. 

In this paper, we study the impact of data processing on the fairness and distortaion of classifiers. We formulate the design of the pre-processor as a convex optimization problem, which for a given possibly discriminating classifier and an acceptable distortion upper bound, aims to minimize a certain measure of discrimination, while satisfying the distortion constraint. We further reformulate a previously proposed post-processing method, which enables us to compare pre and post processing techniques. We show that under some mild assumptions, pre-processing outperforms post-processing.

Existing literature on pre-processing techniques for mitigating disparate impact includes representation learning \cite{xu2018fairgan}, reweighing or resampling the data \cite{kamiran2012data}, and modifying individual records \cite{hajian2013methodology}. For example, in \cite{calmon2017optimized}, the authors focused on the trade-off between discrimination control, utility, and individual distortion. In \cite{feldman2015certifying}, the authors solved an optimal transport problem for designing data transformations. It is worth noting that the design of transformations in both aforementioned approaches requires the use of the sensitive attribute. Using the sensitive attribute in the input of the classifier is a case of disparate treatment, yet this attribute can be used in the pre and post processors. We will consider the effect of feeding the sensitive attribute to the pre-processing module, and will show that having this as input can significantly improve the performance. 

Our work is inspired by recent information-theoretic studies of fairness. For instance, \cite{wang2018direction} derived a correction function to identify proxy variables, which may cause disparate impact. In \cite{ghassami2018fairness}, the authors proposed an information-theoretic pre-processing method to map features to an intermediate variable that is highly informative about the true outcome, while lacks information about the sensitive attribute. However, the resulting optimization problem in \cite{ghassami2018fairness} is not convex, and lacks convergence guarantees. In \cite{wei2019optimized}, the authors propose an optimization formulation for transforming score functions (the predicted probability of being in the positive class by a classifier) to satisfy fairness constraints while minimizing the loss in utility. In \cite{wang2019repairing}, the authors introduce a descent algorithm to perturb the distribution of the input variables of a given classifier to reduce discrimination. Our approach is also related to work on information-theoretic privacy (see e.g., \cite{du2012privacy,sankar2013utility,issa2016operational}), which seeks to characterize the privacy-utility trade-offs using information-theoretic metrics and design privacy-assuring mappings that approach this fundamental trade-off. 

The contributions of this paper are as follows:
\begin{itemize}
    \item We analyze the fundamental trade-off between fairness and accuracy for data pre-processing technique from an information-theoretic approach. We propose a convex program to design a pre-processor, which reduce discrimination, while satisfy a certain accuracy guarantee. 
    \item We characterize the properties of achievable lower bound of discrimination as a function of accuracy for a system with pre-processor.
    \item 
    We reformulate the post-processor proposed in \cite{hardt2016equality} in terms of our fairness and accuracy measures and compare the resulting post-processor with our proposed pre-processor. Under some mild assumptions, we show that any post-processor can be substituted with a pre-processor, while there exist a pre-processor which achieves a better accuracy-fairness combination than any post-processor.
    \item Finally, we show that when the total variation is used to measure discrimination, our optimization problems for designing pre and post-processor will be linear and hence can be solved efficiently. We characterize the achievable fairness-accuracy region using sensitivity analysis of the linear program.
\end{itemize}
The rest of the paper is organized as follows. In Section~\ref{sec:SM} measures of fairness and accuracy are formally described. In Section \ref{sec:design}  we formulate an optimization problem in order to design pre and post-processor. We further characterize the properties of this optimization in this section. In Section \ref{sec:comparison} we compare pre-processing and post-processing. In section \ref{sec:AS} we show that the optimization problem can be reduced to a linear program. 
Our concluding remarks are presented in Section~\ref{sec:con}.

\section{Model Description}\label{sec:SM}

We consider a statistical setting where each individual posseses a set of features, denoted by vector $X\in\mathcal{X}$, and a binary sensitive attribute, denoted by $A$.  The majority group is represented by $A=0$, and the minority group is represented by $A=1$. Each individual also attains a true outcome, which we denote by $Y\in\mathcal{Y}$. Without loss of generality, we assume $\mathcal{Y}=\{0,1,\dots,|\mathcal{Y}|\}$. We assume a classifier (e.g., logistic regression), represented by a transformation $\W$, is given, which takes the  feature vector $X$ as the input and outputs a prediction $\hat{Y}_O$. Although $\W$ does not take $A$ as an input, it might be discriminatory due to dependence between $A$ and $X$. Further we denote our sanitized fair prediction resulting from a system containing a pre-processor and post-processor module by  $\hat{Y}_F$ and $\hat{Y}_P$, respectively. We assume we are given an auditing dataset, which is drawn i.i.d. from the joint distribution $P_{A,X,Y}$. Throughout the paper, we make the common information-theoretic assumption that the joint distribution is given, as it can be estimated from data. We first define our measures for accuracy and fairness. 

\subsection{Accuracy Measure} 
Let $\hat{Y}_O$ be the original outcome of $\W$, and $\hat{Y}$ be our sanitized (by either pre or post processing) fair prediction. To measure accuracy, we assume a distortion function $d(\cdot,\cdot): \mathcal{Y}\times \mathcal{Y}\to \mathbb{R}$, which satisfies $d(y,y)=0~ \forall y$, is given. $\EE{d(Y,\Hat{Y})}$ represents the expected value of the output distortion after data processing. We require distortion of the prediction after data processing to be bounded by a threshold $D$. Formally,
\begin{equation}
\EE{d(Y,\Hat{Y})} \leq D.
\label{eq:dist}
\end{equation}

One can choose a more restrictive distortion constraint by considering
\begin{align}\label{eq:confitional}
\EE{d(Y,\Hat{Y})|X=x} \leq D~~\forall x.
\end{align}

In supplementary material, Section~\ref{sec:AP2} we demonstrate that our results can be derived for the conditional distortion constraint as well.

\subsection{Discrimination Measures} 
In our work, we use equalized odds \cite{hardt2016equality} as our discrimination criterion, which is defined as follows.

\begin{defn} \label{def:eqodds}
Prediction outcome $\hat{Y}$ satisfies equalized odds criterion if for any $\hat{y},y$
\begin{align}\label{eq:EOdds}
\Pr(\hat{Y}=\hat{y}|A=0,Y=y) =  \Pr(\hat{Y}=\hat{y}|A=1,Y=y).
\end{align}
\end{defn}

\begin{rem}
demographic parity \cite{calders2009building} is another widely used fairness criterion. Prediction outcome $\hat{Y}$ satisfies demographic parity criterion if for any $\hat{y}$
\begin{align*}
    \Pr(\hat{Y}=\hat{y}|A=0) = \Pr(\hat{Y}=\hat{y}|A=1).
\end{align*}

Although in the rest of the paper we will use equalized odds as our fairness criterion, in supplementary material, Section \ref{sec:AP3} we will prove that all the properties can be derived similarly for demographic parity as well.
\end{rem}
In order to measure discrimination, we use $f$-divergence to calculate the amount to which the equality in Definition~\ref{def:eqodds} is violated. Hence, we measure the discrimination in the equalized odds sense as,
\begin{align*}
\fdiv\big(P_{\hat{Y}|Y,A=0}\|P_{\hat{Y}|Y,A=1}\big),
\end{align*}
where $\fdiv(\cdot\|\cdot)$ denotes $f$-divergence \cite{ali1966general}.

\begin{rem}
Conditional mutual information between the sensitive attribute and the prediction conditioned on the true outcome  $I(A;\hat{Y}|Y)$ measures the dependency between these two variables. Hence, it might be tempting to use this quantity for quantifying the discrimination\footnote{Similarly one may suggest $I(A;\Hat{Y})$ as the measure of discrimination in the demographic parity case.}. We argue that mutual information is not a suitable measure for quantifying discrimination when the number of samples from majority group is much larger than the minority group. To illustrate this, consider the following factorization.
\begin{align*}
I(A;\hat{Y}|Y)=\sum_y &P_Y(y)\bigg[P_{A|Y}(0|y)\KL(P_{\hat{Y}|Y=y,A=0}\|P_{\hat{Y}|Y=y}) + P_{A|Y}(1|y)\KL(P_{\hat{Y}|Y=y,A=1}\|P_{\hat{Y}|Y=y})\bigg],
\end{align*}

\begin{figure}
    \centering
    \includegraphics[scale=0.32]{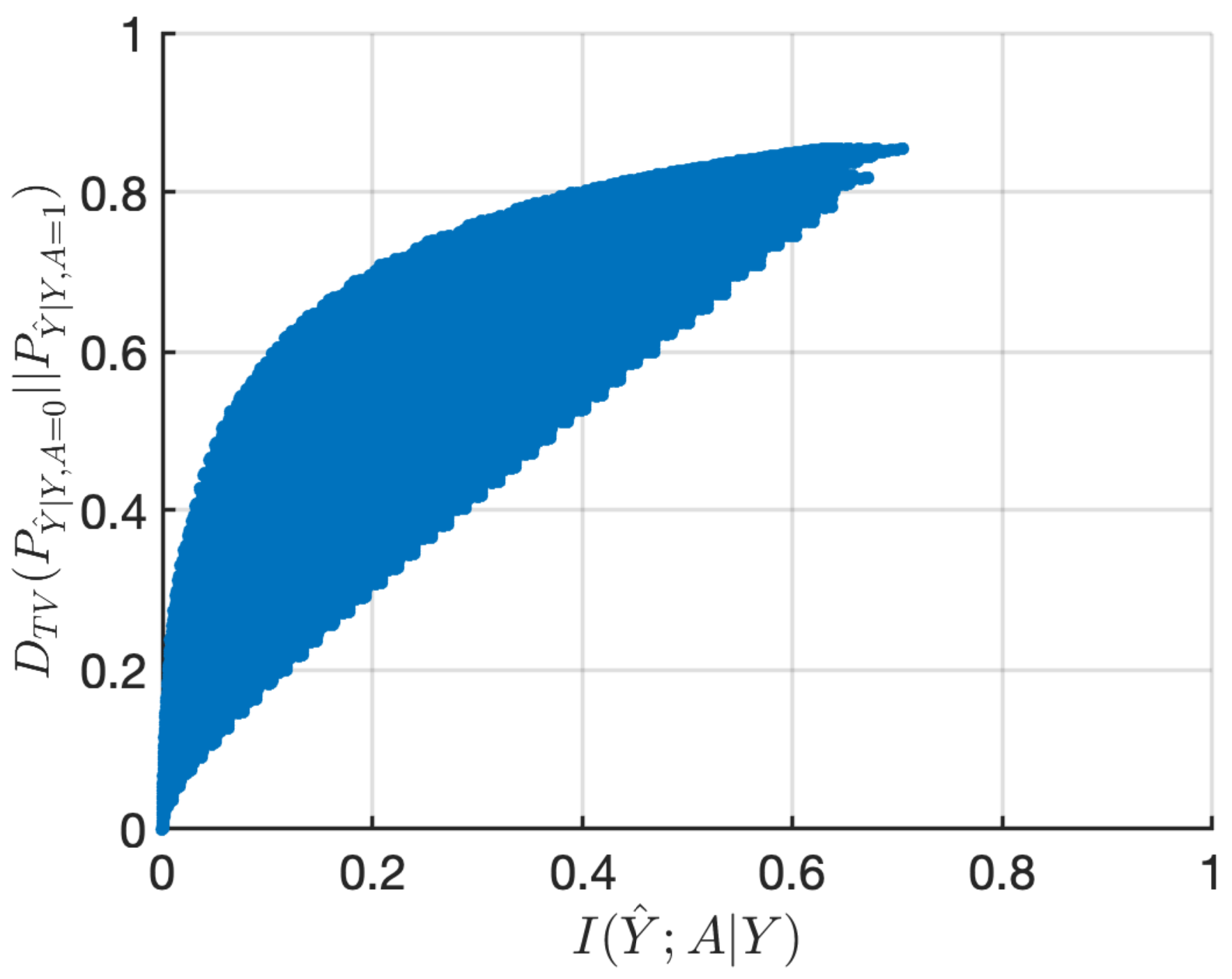}
    \caption{Feasible region in the graph of $\big(D_{TV}(P_{\Hat{Y}|Y,A=0}||P_{\Hat{Y}|Y,A=1})~,~I(\Hat{Y};A|Y)\big)$. The graph is numerically generated by initializing the joint distribution $P_{\Hat{Y},A,Y}$ uniformly at random, and calculating the corresponding point in the graph.}
    \label{fig:2.5}
\end{figure}

where $\KL(\cdot\|\cdot)$ is the KL-divergence. Observe that if almost the entire dataset is from the majority demographic $A=0$, i.e., $\Pr(A=0)\approx1$, then since $P_{\hat{Y}|Y=y,A=0}\approx P_{\hat{Y}|Y=y}$, regardless of the predictor, $I(A,\hat{Y}|Y)\approx0$. Hence, mutual information cannot measure the fairness of the classifier. Figure \ref{fig:2.5} shows the relation between the mutual information definition of equalized odds, $I(A;\hat{Y}|Y)$, and the Total Variation (TV)-distance $\TV(P_{\Hat{Y}|Y,A=0}||P_{\Hat{Y}|Y,A=1})$. It can be seen that for small values of $I(\Hat{Y};A|Y)$, TV-distance can attain large values.
As a result, upper bounding $I(\Hat{Y};A|Y)$, does not result in an upper bound for $\TV(P_{\Hat{Y}|Y,A=0}||P_{\Hat{Y}|Y,A=1})$.
\end{rem}

\section{Designing Data Processor}\label{sec:design}

In this section, our goal is to design a pre-processor $P_{\tilde{X}|X}$ ($P_{\tilde{X}|X,A}$) and a post-processor $P_{\Hat{Y}_P|\Hat{Y}_O,A}$ in order to reduce discrimination. 
\subsection{Pre-processor}
\begin{figure}
    \centering
    \includegraphics[scale=0.35]{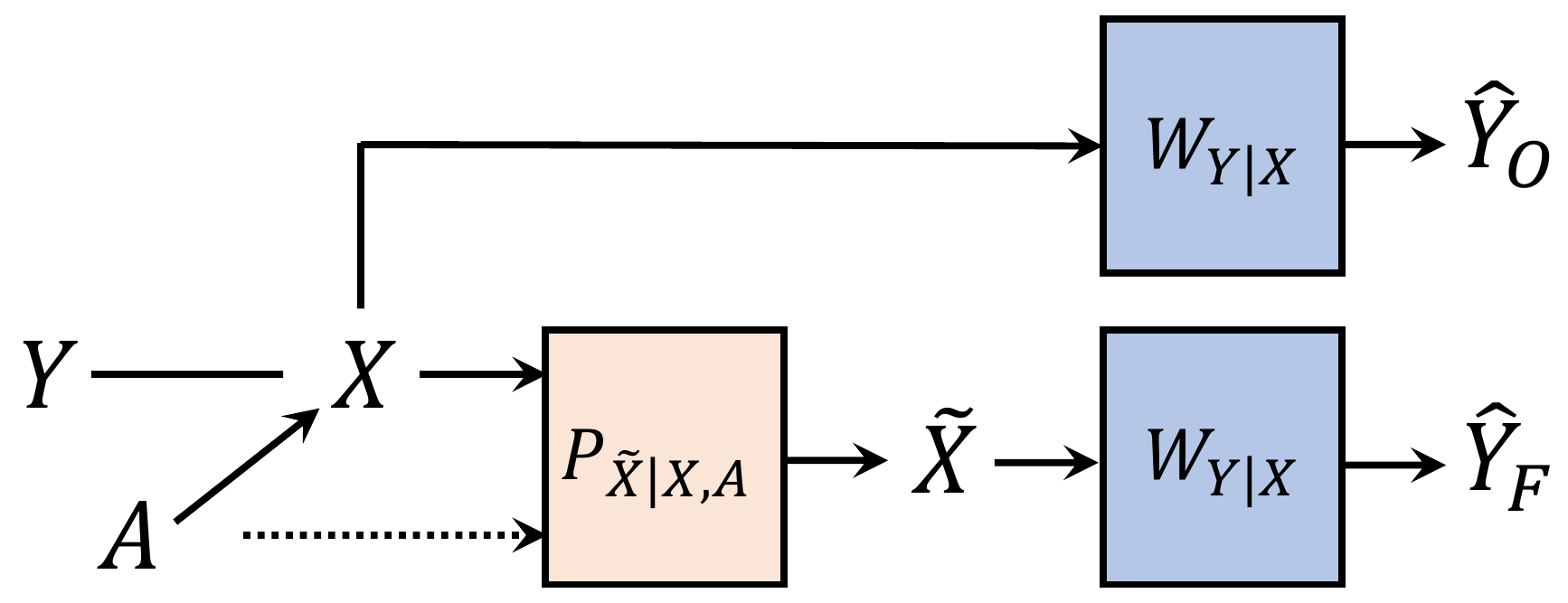}
    \caption{Graphical model of the pre-processing method.}
    \label{fig:GM}
\end{figure}

Figure~\ref{fig:GM} represents the graphical model of the pre-processing setup, in which $\hat{Y}_F$ is the sanitized fair prediction resulting from applying $\W$ to the output of the pre-processed features $\Tilde{X}$ (and $A$).
We require $\hat{Y}_F$ to be  ``as accurate as possible'', and ``as fair as possible''. For a given classifier $W$ and a distortion threshold $D$, the pre-processing fairness-accuracy trade-off function $\mathsf{Disc}^{\mathsf{pre}}_f(W,D)$ is defined as follows.
\begin{align}
        \mathsf{Disc}^{\mathsf{pre}}_f(\W,D)&=
        \min_{P_{\Tilde{X}|X}}\fdiv(P_{\hat{Y}_F|Y,A=0}\|P_{\hat{Y}_F|Y,A=1})\label{eq:opt_one}\\
        &\sto~\EE{d(Y,\Hat{Y}_F)} \leq D. ~~~~\label{eq:opt_three}
\end{align}

$\mathsf{Disc}^{\mathsf{pre}}_f(\W,D)$ provides a fundamental lower bound on discrimination for any pre-processing method. 

In the formulation of equation \eqref{eq:opt_one} and \eqref{eq:opt_three}, only the feature vector $X$ is taken as input to the pre-processing module. One can further input the protected attribute to the pre-processing module, and instead of $P_{\tilde{X}|X}$, optimize over $P_{\tilde{X}|X,A}$. We denote the output of such optimization as $\mathsf{Disc}^{\mathsf{pre|A}}_f(\W,D)$. Since $\mathsf{Disc}^{\mathsf{pre}}$ is a special case of $\mathsf{Disc}^{\mathsf{pre|A}}$, it is expected that we have $\mathsf{Disc}^{\mathsf{pre}}_f(\W,D)\geq \mathsf{Disc}^{\mathsf{pre|A}}_f(\W,D)$, for all $D$. 

\subsection{Post-processor}
Given a possibly biased classifier $W_{Y|X}$, the authors in \cite{hardt2016equality} proposed a post-processing method for finding a post hoc correction, $\hat{Y}_P$ to the output $\Hat{Y}_O$ of the classifier $W$. Given the joint distribution $P_{\Hat{Y}_O, A, Y}$, the authors propose the following optimization problem.
\begin{align} \label{eq:hardt}
\begin{split}
    &\min_{P_{\hat{Y}_P|\Hat{Y}_O,A}}~~ E[d(Y,\hat{Y}_P)]\\
    &\text{s.t.} ~~ \Pr(\hat{Y}_P=\hat{y}_P|A=0,Y=y)   
    =\Pr(\hat{Y}_P=\hat{y}_P|A=1,Y=y),\forall y,\hat{y}_P.
\end{split}
\end{align}

\begin{figure}
    \centering
    \includegraphics[scale=0.35]{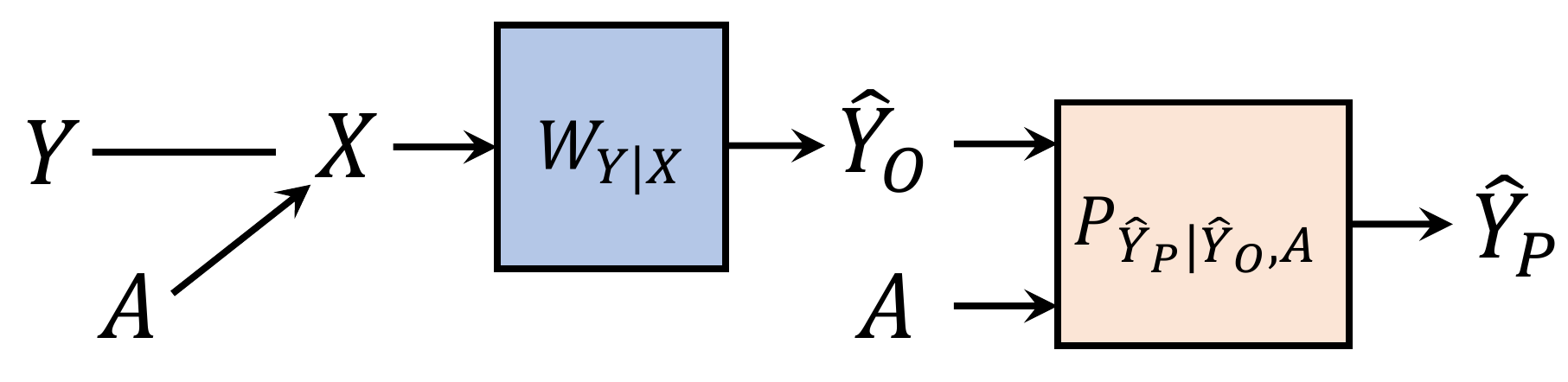}
    \caption{Graphical model of the post-processing method.}
    \label{fig:GM2}
\end{figure}

However, as stated in \cite{woodworth2017learning}, satisfying exact equalized odds when dealing with finite data set may result in trivial predictor $\hat{Y}_P=1$ or $\hat{Y}_P=0$. We propose the following optimization formulation, where $\mathsf{Disc}^{\mathsf{post}}_f(\W,D)$ gives the lowest attainable discrimination (in the equalized odds sense) via post-processing the output of the classifier $W$, when distortion is upper bounded with $D$. Figure \ref{fig:GM2} represents the graphical model of the post-processing method. 

\begin{align}
    \begin{split}\label{eq:opt_four}
        \mathsf{Disc}^{\mathsf{post}}_f(\W,D)&=\min_{P_{\Hat{Y}_P|\Hat{Y}_O,A}}\fdiv(P_{\hat{Y}_P|Y,A=0}||P_{\hat{Y}_P|Y,A=1}),
    \end{split}\\
    \begin{split}\label{eq:opt_five}
        \sto~\EE{d(Y,\Hat{Y}_P)} \leq D. ~~~~
    \end{split}
\end{align}

In this formulation, instead of requiring the equality constraint in equation \eqref{eq:EOdds}, we minimize discrimination conditioned on an upper bound on the distortion. Since we are not constraining the output to satisfy exact fairness, in the case of finite data, the post-processing module will not be forced to generate a low accuracy output just to satisfy equation \eqref{eq:EOdds} with equality. In addition, in this formulation we have a tunable hyperparameter $D$ that can be used to trade fairness for accuracy and vice versa.

\subsection{Properties of Fair Data Processing} 
\label{sec:PD}

In this section, we show that the optimization programs presented in \eqref{eq:opt_one} and \eqref{eq:opt_four} are convex, and we characterize the properties of $\mathsf{Disc}^{\mathsf{pre}}_f$ and $\mathsf{Disc}^{\mathsf{post}}_f$.
We first observe some properties regarding the constraints in the optimization problems.
\begin{lem}\label{lem:trick}
(a) Distortion constraint in \eqref{eq:opt_three} can be written as 
\begin{align*}
    \EE{\bar{d}(\tilde{X},X)} \leq D,
\end{align*}
where $\bar{d}(\tilde{x},x) = \sum_{y,\hat{y}_F}W_{Y|X}(\hat{y}_F|\tilde{x})P_{Y|X}(y|x)d(y,\hat{y}_F)$.

(b) The constraints in \eqref{eq:opt_three} and \eqref{eq:opt_five} are linear with respect to $P_{\tilde{X}|X}$ and $P_{\hat{Y}_P|\hat{Y}_O,A}$, respectively.
\end{lem}
See supplementary material, Section~\ref{sec:AP-1} for a proof.

As stated earlier, since the optimization problem of the proposed pre-processing method  in \cite{ghassami2018fairness} is not convex, the authors could not provide any convergence guarantees for their method. In the following theorem, we observe that our formulation has the convexity property.

\begin{prop}\label{prop:con}
Both of the optimization problems in \eqref{eq:opt_one}-\eqref{eq:opt_three} and \eqref{eq:opt_four}-\eqref{eq:opt_five} are convex programs.
\end{prop}

See supplementary material, Section~\ref{sec:AP0} for a proof.

Since the problem is convex, standard convex optimization techniques can be used to find the global minimum. 

The following theorem describes the characteristics of $\mathsf{Disc}^{\mathsf{pre|A}}_f(W,D)$ and $\mathsf{Disc}^{\mathsf{post}}_f(W,D)$ functions.
\begin{prop}\label{prop:1}
For a given $W$ and the joint distribution $P_{A,X,Y}$, the mappings $D\mapsto \mathsf{Disc}^{\mathsf{pre|A}}_f(\W,D)$ and $D\mapsto \mathsf{Disc}^{\mathsf{post}}_f(\W,D)$ satisfy the following properties:

\begin{enumerate}
    \item There exist $D'$, such that $\mathsf{Disc}^{\mathsf{pre|A}}_f(\W,D)=0$ for $D\geq D'$. We denote smallest such $D'$ by $D_{\max}^{\mathsf{pre|A}}$. We have the following bound:
    \begin{align*}
    D_{\max}^{\mathsf{pre|A}} \leq& \frac{1}{|\mathcal{X}|} \sum_{\tilde{x},y,\Hat{y}_F}W_{Y|X}(\Hat{y}_F|\tilde{x})P_Y(y)d(y,\Hat{y}_F) 
    \end{align*}
    Similarly, there exist $D_{\max}^{\mathsf{post}}$, such that $\mathsf{Disc}^{\mathsf{post}}_f(\W,D)=0$ for $D\geq D_{\max}^{\mathsf{post}}$.
    \item There exist $D_{\min}^{\mathsf{pre|A}}$, such that the constraint in \eqref{eq:opt_three} is infeasible for $D<D_{\min}^{\mathsf{pre|A}}$, and feasible for all $D\geq D_{\min}^{\mathsf{pre|A}}$. $D_{\min}^{\mathsf{pre|A}}$ is given by:
    \[
    D_{\min}^{\mathsf{pre|A}} = \sum_{x,y,\Hat{y}_F,a} W_{Y|X}(\Hat{y}_F|\tilde{x}_{(x,a)}) P_{X,Y,A}(x,y,a)d(y,\Hat{y}_F),
    \]
    where
    \[
    \tilde{x}_{(x,a)} = \argmin_{\tilde{x}} \sum_{y,\Hat{y}_F}  W_{Y|X}(\Hat{y}_F|\tilde{x})P_{X,Y,A}(x,y,a)d(y,\Hat{y}_F).
    \]
    The pre-processor $P_{\tilde{X}|X,A}$ corresponding to $\mathsf{Disc}_{f}^{\mathsf{pre|A}}(W,D_{\min})$ is given by
    \[
    P_{\tilde{X}|X,A}(\tilde{x}|x,a) = 
    \begin{cases}
         1, &  \forall x, \tilde{x} = \tilde{x}_{(x,a)},\\    
         0, & \text{otherwise.}  
    \end{cases}
    \]
    Similarly, there exist $D_{\min}^{\mathsf{post}}$, such that the constraint is \eqref{eq:opt_five} is infeasible for all $D<D_{\min}^{\mathsf{post}}$.
    \item $\mathsf{Disc}^{\mathsf{pre|A}}_f(\W,D)$ and $\mathsf{Disc}^{\mathsf{post}}_f(\W,D)$ are both convex with respect to $D$. 
    \item Suppose \eqref{eq:opt_one} has a positive minimum for some $D<D_{\max}^{\mathsf{pre|A}}$. For any $D\leq D_{\max}^{\mathsf{pre|A}}$, $\mathsf{Disc}^{\mathsf{pre|A}}_f(\W,D)$ is strictly decreasing in $D$. The same property holds for $\mathsf{Disc}^{\mathsf{post}}_f(\W,D)$ for $D\leq D_{\max}^{\mathsf{post}}$.
    \item We have $\mathsf{Disc}^{\mathsf{pre|A}}_f(\W,D_{\min}^{\mathsf{pre}|A}) \leq \fdiv(P_{\hat{Y}_O|Y,A=0}||P_{\hat{Y}_O|Y,A=1})$. 
\end{enumerate}
\end{prop}
See supplementary material, Section \ref{sec:Ap1} for the proof.

\begin{rem}\label{rem:Dneg}
As mentioned earlier, the constraint in \eqref{eq:opt_three} controls the extra distortion due to the pre-processing module. At first glance, one may anticipate that the problem should be infeasible for $D<\EE{d(Y,\hat{Y}_O)}$ (i.e., due to data processing inequality insights, it is impossible to reduce the distortion by adding the pre-processing module). However, $\W$ could be the result of any classification algorithm, such as Logistic Regression, Decision Trees, etc. Since in all such algorithms the set of achievable classifiers $\W$ is limited, a transformation on the input may lead to better performance of the designed classifier, and hence, the problem can be feasible for $D<\EE{d(Y,\hat{Y}_O)}$. 
\end{rem}

\section{Comparison of Pre and Post Processing Methods}\label{sec:comparison}
In this section, we provide a theoretical comparison of the proposed pre-processing and post-processing methods. 
Specifically,
we show that in general, pre-processing is more powerful in the sense that under some mild conditions a post-processor can always be substituted by a pre-processor, while there exist pre-processors, which cannot be outperformed by any post-processor. In all the results, we consider a binary label, i.e., $|\mathcal{Y}|=2$.
The following theorem provides a necessary and sufficient condition for replacing a post-processor with a pre-processor.

\begin{prop}\label{prop:substitution}
Any post-processer can be substituted with a pre-processor if and only if there exist $x_0,x_1 \in \mathcal{X}$ such that
\[
W_{Y|X}(1|x_0) = 0, ~~ W_{Y|X}(1|x_1) = 1.
\]
\end{prop}
See supplementary materials, Section \ref{sec:nonexistence} for a proof.

Proposition \ref{prop:substitution} states that if there exist two individuals $x_0, x_1\in\mathcal{X}$ that the classifier $W$  classify deterministically as, say, unqualified and qualified, respectively, then any post-processing module designed for $W$ can be substituted with a pre-processing module which results in the same level of distortion and discrimination as the post-processor. Since the cardinality of the feature space $|\mathcal{X}|$ is usually relatively large, it is expected that the requirement of Proposition \ref{prop:substitution} is satisfied for most of the common classifiers.

\begin{defn}
A predictor $W_{Y|X}$ is called proper if for $a\in\{0,1\}$, we have

\[
P_{\hat{Y}_O,Y,A}(1,1,a) > P_{\hat{Y}_O,Y,A}(\hat{y}_O,y,a),  ~~ \hat{y}_O\neq y,
\]
and
\[
P_{\hat{Y}_O,Y,A}(0,0,a) > P_{\hat{Y}_O,Y,A}(\hat{y}_O,y,a), ~~ \hat{y}_O\neq y.
\]
\end{defn}

Let $Y=1$ represent the label corresponding to the desired property (e.g., being qualified in the task of hiring). As mentioned earlier, we represent the minority demographic by $A=1$. Since the minority group is the underprivileged demographic, we assume that  $P_{\hat{Y}_O|Y,A}(1|1,0) > P_{\hat{Y}_O|Y,A}(1|1,1)$ and $P_{\hat{Y}_O|Y,A}(1|0,0) > P_{\hat{Y}_O|Y,A}(1|0,1)$. 
Recall that the minimum feasible prediction distortion for pre and post processing is denoted by $D_{\min}^{\mathsf{pre}}$ and $D_{\min}^{\mathsf{post}}$, respectively. Proposition \ref{prop:10} states the condition under which there exist a pre-processor that has lower disortion than any post-processor, while has lower discrimination than $\mathsf{Disc}_f^{\mathsf{post}}(W,D_{\min}^{\mathsf{post}})$. In this proposition, we assume $x_{\min}=\argmin_{x} W_{Y|X}(1|x)$, and similarly for $x_{\max}$.

\begin{prop}\label{prop:10}
Given a proper predictor $W$, if there exist $x_i\neq x_{\max}$, such that $P_{Y|X,A} (0|x_i,1) < P_{Y|X,A} (1|x_i,1)$, or there exist $x_j\neq x_{\min}$, such that $P_{Y|X,A} (1|x_j,0) < P_{Y|X,A} (0|x_j,0)$, then we have $D_{\min}^{\mathsf{pre|A}} < D_{\min}^{\mathsf{post}}$, and $\mathsf{Disc}_f^{\mathsf{pre|A}}(W,D_{\min}^{\mathsf{pre|A}})< \mathsf{Disc}_f^{\mathsf{post}}(W,D_{\min}^{\mathsf{post}})$.
\end{prop}
See supplementary material, Section \ref{sec:AP20} for the proof.

The condition in this proposition is arguably mild. The condition regarding $x_i$ is satisfied simply if there exist an individual $x_i$ in the minority group, which is qualified with probability larger than $1/2$, and the condition regarding $x_j$ is satisfied simply if there exist an individual $x_j$ in the majority group, which is unqualified with probability larger than $1/2$.

\section{Linear Programming Formulation} \label{sec:AS}
In the previous section, we proved that the optimization problems presented in ~\eqref{eq:opt_one}-\eqref{eq:opt_three} and \eqref{eq:opt_four}-\eqref{eq:opt_five} are both convex for any $f$-divergence. Hence, one can use any standard numerical convex optimization approach to solve the problem. However, when the cardinality of features (i.e., $|\mathcal{X}|$) is large, the numerical solution of the proposed optimization in \eqref{eq:opt_one}-\eqref{eq:opt_three} can be computationally infeasible. Consequently, we propose a linear programming solution to the problem when total variation is used as a special case of $f$-divergence. We note that total variation is symmetric with respect to its input distributions, which is a desired property in our setting; a property which is not satisfied by some other candidates such as KL-divergence.

In order to solve the optimization problem, we first extend the optimization argument to $P_{\tilde{X}|X}$ and $P_{\hat{Y}_F|A}$ and add the constraint $P_{\hat{Y}_F|Y,A}(\hat{y}_F|y,a)=\sum_{x,\tilde{x}} \W(\Hat{y}_F|\tilde{x})P_{\hat{X}|X}(\tilde{x}|x)P_{X|Y,A}(x|y,a)$, which is justified due to the graphical model in Figure~\ref{fig:GM}. Note that total variation is of the form of a summation over absolute values and it is well known that (see, e.g., \cite{bertsimas1997introduction}) the optimization problem
\begin{align*}
&\min_{z_1,\dots, z_n} \sum_i |z_i|,   \\
&~~\text{s.t.}~~ f(z_1,\dots,z_n) = 0,
\end{align*}
is equivalent to
\begin{align*}
\min_{z_1,\dots, z_n, t_1,\dots,t_n} &\sum_i t_i,\\
\text{s.t.}~~~& t_i \geq z_i, \\
 &t_i \geq -z_i,\\
 &f(z_1,\dots,z_n) = 0.
\end{align*}
Therefore, introducing variables $t_{y,\hat{y}_F},$ for $y,\hat{y}_F \in \mathcal{Y}$, the optimization problem in \eqref{eq:opt_one}-\eqref{eq:opt_three} can be written as follows.
\begin{equation}\label{eq:LP}
\begin{split}
 \mathsf{Disc}^{\mathsf{pre|A}}_{TV}(W,D)&=
 \min_{P_{\Tilde{X}|X},P_{\Hat{Y}_F|Y,A},t}~~ \sum_{y=1}^{|\mathcal{Y}|}P_Y(y)\sum_{\hat{y}_F=1}^{|\mathcal{Y}|} t_{y,\hat{y}_F},\\
 \text{s.t.}&~~P_{\Hat{Y}_F|Y,A}(\hat{y}_F|y,0)-P_{\Hat{Y}_F|Y,A}(\hat{y}_F|y,1)  - t_{y,\hat{y}_F}\leq 0, ~~\forall y,\Hat{y}_F,\\
&P_{\Hat{Y}_F|Y,A}(\hat{y}_F|y,1)-P_{\Hat{Y}_F|Y,A}(\hat{y}_F|y,0)- t_{y,\hat{y}_F}\leq 0, ~~\forall y,\Hat{y}_F,\\
 &\sum_{\substack{x,\tilde{x},y, \\ \hat{y}_F,a} }  W_{Y|X}(\hat{y}_F|\tilde{x})P_{\tilde{X}|X,A}(\tilde{x}|x,a)P_{X,Y,A}(x,y,a)d(y,\hat{y}_F) \leq  D\\
 &P_{\Hat{Y}_F|Y,A}(\Hat{y}_F|y,a) ~\!=\!\sum_{x,\tilde{x}} W_{Y|X}(\Hat{y}_F|\tilde{x})P_{\tilde{X}|X,A}(\tilde{x}|x,a)P_{X|Y,A}(x|y,a), \forall \Hat{y}_F,y,a\\
 &\sum_{\hat{y}_F} P_{\Hat{Y}_F|Y,A}(\hat{y}_F|y,a) = 1,~P_{\Hat{Y}_F|Y,A}(\hat{y}_F|y,a) \geq 0,~ \forall \hat{y}_F,y, a,  \\
 &\sum_{\tilde{x}} P_{\tilde{X}|X,A}(\tilde{x}|x,a) = 1, P_{\tilde{X}|X,A}(\tilde{x}|x,a)  \geq 0, \forall\tilde{x},x,a,
\end{split}
\end{equation}
which is a linear program and can be solved efficiently in polynomial time. 

similarly, using TV-distance, the optimization in \eqref{eq:opt_four}-\eqref{eq:opt_five} reduces to a linear program. The Resulting form is presented in supplementary material, Section \ref{sec:AP40}.

One can further analyze the properties of $\mathsf{Disc}_{TV}^{\mathsf{pre}}(W,D)$ and $\mathsf{Disc}_{TV}^{\mathsf{post}}(W,D)$ as a function of $D$. We have the following result in this regard.

\begin{lem}\label{lem:poly}
$D\mapsto \mathsf{Disc}^{\mathsf{pre|A}}_{TV}(\W,D)$ and $D\mapsto \mathsf{Disc}^{\mathsf{post}}_{TV}(\W,D)$ are piecewise linear functions.
\end{lem}
See supplementary material, Section \ref{sec:AP5} for the proof.

\section{Conclusion}\label{sec:con}
In this paper we analysed pre and post processing methods for reducing discrimination. We proposed an optimizations problem which results in a pre-processing module that can be added before a classifier, and reduce prediction discrimination, while ensures a distortion upper bound in the output. We proved that our optimization is convex, hence the global minimum is achievable. We reformulated an already proposed post-processing method as a convex optimization. Furthermore, we compared pre and post processing methods, and we showed under some mild assumptions pre-processing outperforms post-processing. Finally, we showed that for a special case of discrimination measure, the optimization problem reduces to a linear program and can be solved efficiently in polynomial time.

\bibliographystyle{IEEEtran}
\bibliography{refrences}

\begin{thebibliography}{10}
\providecommand{\url}[1]{#1}
\csname url@samestyle\endcsname
\providecommand{\newblock}{\relax}
\providecommand{\bibinfo}[2]{#2}
\providecommand{\BIBentrySTDinterwordspacing}{\spaceskip=0pt\relax}
\providecommand{\BIBentryALTinterwordstretchfactor}{4}
\providecommand{\BIBentryALTinterwordspacing}{\spaceskip=\fontdimen2\font plus
\BIBentryALTinterwordstretchfactor\fontdimen3\font minus
  \fontdimen4\font\relax}
\providecommand{\BIBforeignlanguage}[2]{{%
\expandafter\ifx\csname l@#1\endcsname\relax
\typeout{** WARNING: IEEEtran.bst: No hyphenation pattern has been}%
\typeout{** loaded for the language `#1'. Using the pattern for}%
\typeout{** the default language instead.}%
\else
\language=\csname l@#1\endcsname
\fi
#2}}
\providecommand{\BIBdecl}{\relax}
\BIBdecl

\bibitem{kinyanjui2019estimating}
N.~M. Kinyanjui, T.~Odonga, C.~Cintas, N.~C. Codella, R.~Panda, P.~Sattigeri,
  and K.~R. Varshney, ``Estimating skin tone and effects on classification
  performance in dermatology datasets,'' \emph{arXiv preprint
  arXiv:1910.13268}, 2019.

\bibitem{buolamwini2018gender}
J.~Buolamwini and T.~Gebru, ``Gender shades: Intersectional accuracy
  disparities in commercial gender classification,'' in \emph{Conference on
  Fairness, Accountability and Transparency}, 2018.

\bibitem{tan2017detecting}
S.~Tan, R.~Caruana, G.~Hooker, and Y.~Lou, ``Detecting bias in black-box models
  using transparent model distillation,'' \emph{Artificial Intelligence, Ethics
  and Society}, 2017.

\bibitem{mendez1979presumptions}
M.~A. Mendez, ``Presumptions of discriminatory motive in title vii disparate
  treatment cases,'' \emph{Stan. L. Rev.}, vol.~32, p. 1129, 1979.

\bibitem{barocas2016big}
S.~Barocas and A.~D. Selbst, ``Big data's disparate impact,'' \emph{Cal. L.
  Rev.}, 2016.

\bibitem{feldman2015certifying}
M.~Feldman, S.~A. Friedler, J.~Moeller, C.~Scheidegger, and
  S.~Venkatasubramanian, ``Certifying and removing disparate impact,'' in
  \emph{Proceedings of the 21th ACM SIGKDD International Conference on
  Knowledge Discovery and Data Mining}.\hskip 1em plus 0.5em minus 0.4em\relax
  ACM, 2015, pp. 259--268.

\bibitem{wang2020split}
H.~Wang, H.~Hsu, M.~Diaz, and F.~P. Calmon, ``To split or not to split: The
  impact of disparate treatment in classification,'' \emph{arXiv preprint
  arXiv:2002.04788}, 2020.

\bibitem{hardt2016equality}
M.~Hardt, E.~Price, N.~Srebro \emph{et~al.}, ``Equality of opportunity in
  supervised learning,'' in \emph{Advances in neural information processing
  systems}, 2016.

\bibitem{berk2018fairness}
R.~Berk, H.~Heidari, S.~Jabbari, M.~Kearns, and A.~Roth, ``Fairness in criminal
  justice risk assessments: The state of the art,'' \emph{Sociological Methods
  \& Research}, p. 0049124118782533, 2018.

\bibitem{zafar2017parity}
M.~B. Zafar, I.~Valera, M.~Rodriguez, K.~Gummadi, and A.~Weller, ``From parity
  to preference-based notions of fairness in classification,'' in
  \emph{Advances in Neural Information Processing Systems}, 2017, pp. 229--239.

\bibitem{corbett2017algorithmic}
S.~Corbett-Davies, E.~Pierson, A.~Feller, S.~Goel, and A.~Huq, ``Algorithmic
  decision making and the cost of fairness,'' in \emph{Proceedings of the 23rd
  ACM SIGKDD International Conference on Knowledge Discovery and Data
  Mining}.\hskip 1em plus 0.5em minus 0.4em\relax ACM, 2017.

\bibitem{agarwal2018reductions}
A.~Agarwal, A.~Beygelzimer, M.~Dud{\'\i}k, J.~Langford, and H.~Wallach, ``A
  reductions approach to fair classification,'' in \emph{International
  Conference on Machine Learning}.\hskip 1em plus 0.5em minus 0.4em\relax PMLR,
  2018, pp. 60--69.

\bibitem{kamiran2012decision}
F.~Kamiran, A.~Karim, and X.~Zhang, ``Decision theory for discrimination-aware
  classification,'' in \emph{Data Mining (ICDM), 2012 IEEE 12th International
  Conference on}.\hskip 1em plus 0.5em minus 0.4em\relax IEEE, 2012, pp.
  924--929.

\bibitem{zhang2018mitigating}
B.~H. Zhang, B.~Lemoine, and M.~Mitchell, ``Mitigating unwanted biases with
  adversarial learning,'' in \emph{Proceedings of the 2018 AAAI/ACM Conference
  on AI, Ethics, and Society}, 2018, pp. 335--340.

\bibitem{calders2010three}
T.~Calders and S.~Verwer, ``Three naive bayes approaches for
  discrimination-free classification,'' \emph{Data Mining and Knowledge
  Discovery}, vol.~21, no.~2, pp. 277--292, 2010.

\bibitem{celis2019classification}
L.~E. Celis, L.~Huang, V.~Keswani, and N.~K. Vishnoi, ``Classification with
  fairness constraints: A meta-algorithm with provable guarantees,'' in
  \emph{Proceedings of the conference on fairness, accountability, and
  transparency}, 2019, pp. 319--328.

\bibitem{pedreschi2009measuring}
D.~Pedreschi, S.~Ruggieri, and F.~Turini, ``Measuring discrimination in
  socially-sensitive decision records,'' in \emph{Proceedings of the SIAM
  International Conference on Data Mining}.\hskip 1em plus 0.5em minus
  0.4em\relax SIAM, 2009.

\bibitem{wei2019optimized}
D.~Wei, K.~N. Ramamurthy, and F.~d.~P. Calmon, ``Optimized score transformation
  for fair classification,'' \emph{arXiv preprint arXiv:1906.00066}, 2019.

\bibitem{romei2014multidisciplinary}
A.~Romei and S.~Ruggieri, ``A multidisciplinary survey on discrimination
  analysis,'' \emph{The Knowledge Engineering Review}, 2014.

\bibitem{xu2018fairgan}
D.~Xu, S.~Yuan, L.~Zhang, and X.~Wu, ``Fairgan: Fairness-aware generative
  adversarial networks,'' in \emph{2018 IEEE International Conference on Big
  Data (Big Data)}.\hskip 1em plus 0.5em minus 0.4em\relax IEEE, 2018, pp.
  570--575.

\bibitem{kamiran2012data}
F.~Kamiran and T.~Calders, ``Data preprocessing techniques for classification
  without discrimination,'' \emph{Knowledge and Information Systems}, vol.~33,
  no.~1, pp. 1--33, 2012.

\bibitem{hajian2013methodology}
S.~Hajian and J.~Domingo-Ferrer, ``A methodology for direct and indirect
  discrimination prevention in data mining,'' \emph{IEEE transactions on
  knowledge and data engineering}, 2013.

\bibitem{calmon2017optimized}
F.~Calmon, D.~Wei, B.~Vinzamuri, K.~N. Ramamurthy, and K.~R. Varshney,
  ``Optimized pre-processing for discrimination prevention,'' in \emph{Advances
  in Neural Information Processing Systems}, 2017, pp. 3992--4001.

\bibitem{wang2018direction}
H.~Wang, B.~Ustun, and F.~P. Calmon, ``On the direction of discrimination: An
  information-theoretic analysis of disparate impact in machine learning,'' in
  \emph{2018 IEEE International Symposium on Information Theory (ISIT)}.\hskip
  1em plus 0.5em minus 0.4em\relax IEEE, 2018.

\bibitem{ghassami2018fairness}
A.~Ghassami, S.~Khodadadian, and N.~Kiyavash, ``Fairness in supervised
  learning: An information theoretic approach,'' in \emph{2018 IEEE
  International Symposium on Information Theory (ISIT)}.\hskip 1em plus 0.5em
  minus 0.4em\relax IEEE, 2018.

\bibitem{wang2019repairing}
H.~Wang, B.~Ustun, and F.~Calmon, ``Repairing without retraining: Avoiding
  disparate impact with counterfactual distributions,'' in \emph{International
  Conference on Machine Learning}.\hskip 1em plus 0.5em minus 0.4em\relax PMLR,
  2019, pp. 6618--6627.

\bibitem{du2012privacy}
F.~P.~Calmon and N.~Fawaz, ``Privacy against statistical inference,'' in
  \emph{Proc. 50th Annual Allerton Conference on Communication, Control, and
  Computing}.\hskip 1em plus 0.5em minus 0.4em\relax IEEE, 2012, pp.
  1401--1408.

\bibitem{sankar2013utility}
L.~Sankar, S.~R. Rajagopalan, and H.~V. Poor, ``Utility-privacy tradeoffs in
  databases: An information-theoretic approach,'' \emph{IEEE Transactions on
  Information Forensics and Security}, vol.~8, no.~6, pp. 838--852, 2013.

\bibitem{issa2016operational}
I.~Issa, S.~Kamath, and A.~B. Wagner, ``An operational measure of information
  leakage,'' in \emph{Information Science and Systems (CISS), 2016 Annual
  Conference on}.\hskip 1em plus 0.5em minus 0.4em\relax IEEE, 2016, pp.
  234--239.

\bibitem{calders2009building}
T.~Calders, F.~Kamiran, and M.~Pechenizkiy, ``Building classifiers with
  independency constraints,'' in \emph{Data mining workshops, 2009. ICDMW'09.
  IEEE international conference on}.\hskip 1em plus 0.5em minus 0.4em\relax
  IEEE, 2009, pp. 13--18.

\bibitem{ali1966general}
S.~M. Ali and S.~D. Silvey, ``A general class of coefficients of divergence of
  one distribution from another,'' \emph{Journal of the Royal Statistical
  Society. Series B (Methodological)}, pp. 131--142, 1966.

\bibitem{woodworth2017learning}
B.~Woodworth, S.~Gunasekar, M.~I. Ohannessian, and N.~Srebro, ``Learning
  non-discriminatory predictors,'' \emph{arXiv preprint arXiv:1702.06081},
  2017.

\bibitem{bertsimas1997introduction}
D.~Bertsimas and J.~N. Tsitsiklis, \emph{Introduction to linear
  optimization}.\hskip 1em plus 0.5em minus 0.4em\relax Athena Scientific
  Belmont, MA, 1997, vol.~6.

\bibitem{csiszar2004information}
I.~Csisz{\'a}r, P.~C. Shields \emph{et~al.}, ``Information theory and
  statistics: A tutorial,'' \emph{Foundations and Trends{\textregistered} in
  Communications and Information Theory}, vol.~1, no.~4, pp. 417--528, 2004.

\end{thebibliography}

\newpage
\appendix
\appendixpage    

\section{Distortion Constraint Reformation}\label{sec:AP-1}
We have:
\begin{align*}
\EE{d(Y,\hat{Y}_F)}=&\sum_{y,\hat{y}_F}P_{Y,\Hat{Y}_F}(y,\hat{y}_F)d(y,\hat{y}_F)\\
=& \sum_{x,\tilde{x},y,\hat{y}_F} P_{\tilde{X}|X}(\tilde{x}|x)W_{Y|X}(\hat{y}_F|\tilde{x})P_{X,Y}(x,y)d(y,\hat{y}_F)\\
 =& \sum_{x,\tilde{x}}  P_{\tilde{X},X}(\tilde{x},x)\bar{d}(\tilde{x},x)\\
 =& \EE{\bar{d}(\tilde{X},X)},
\end{align*}
where
\[
\bar{d}(\tilde{x},x) = \sum_{y,\hat{y}_F}W_{Y|X}(\hat{y}_F|\tilde{x})P_{Y|X}(y|x)d(y,\hat{y}_F),
\]
which shows that $\EE{d(Y,\hat{Y}_F)}\leq D$ is a linear constraint with respect to the pre-processing channel $P_{\tilde{X}|X,A}$.

Additionally, 
\begin{align*}
    \EE{d(Y,\hat{Y}_P)}=&\sum_{\Hat{y}_P,\Hat{y}_O,y,a} \big[P_{\Hat{Y}_P|\Hat{Y}_O,A}(\Hat{y}_P|\Hat{y}_O,a) P_{\Hat{Y}_O|Y,A}(\Hat{y}_O|y,a)\times P_{Y,A}(y,a)d(y,\Hat{y}_P)\big]
\end{align*}
which is linear with respect to the post-processing channel $P_{\Hat{Y}_P|\Hat{Y}_O,A}$. 

\section{Proof of Convexity}\label{sec:AP0}
\textbf{Convexity of the pre-processing formulation:} First we prove the convexity of the objective function. Note that $f$-Divergence is convex with respect to the joint components \cite{csiszar2004information}:
\[
\fdiv(P||Q) \leq\lambda \fdiv(P_1||Q_1) + (1-\lambda) \fdiv(P_2||Q_2),
\]
where $P=\lambda P_1 + (1-\lambda)P_2$ and $Q=\lambda Q_1 + (1-\lambda)Q_2$ and $0\leq \lambda \leq 1$. Furthermore, due to the graphical model in Figure~\ref{fig:GM}, $P_{\hat{Y}_F|Y,A}$ can be written as a linear function of $P_{\tilde{X}|X}$:
\[
 P_{\hat{Y}_F|Y,A}(\Hat{y}_F|y,a) = \sum_{x,\tilde{x}} W_{Y|X}(\hat{y}_F|\tilde{x}) P_{\tilde{X}|X}(\tilde{x}|x)P_{X|Y,A}(x|y,a).
\]
As a result, $\fdiv(P(\hat{Y}_F|Y,A=0)||P(\hat{Y}_F|Y,A=1))$ is a convex function with respect to $P(\tilde{X}|X)$.

In addition, according to Lemma \ref{lem:trick}, one can write the distortion constraint \eqref{eq:opt_three} as a linear function of $P(\tilde{X}|X)$. Therefore, the optimization problem in \eqref{eq:opt_one} and \eqref{eq:opt_three} is convex.

\textbf{Convexity of the post-processing formulation:} 
Similarly, we have
\[
P_{\Hat{Y}_P|Y,A}(\Hat{y}_P|y,a) = \sum_{\Hat{y}_O,x} P_{\Hat{Y}_P|\Hat{Y}_O,A}(\Hat{y}_P|\Hat{y}_O,a)W_{Y|X}(\Hat{y}_O|x)P_{X|Y,A}(x|y,a),
\]
which is linear with respect to the post-processing channel $P_{\Hat{Y}_P|\Hat{Y}_O,A}$. Again using the convexity of the f-Divergence with respect to the distributions, we can prove that the objective function of the post-processing is convex. In addition, the constraint is linear, which proves the convexity of the post-processing formulation.

\section{Proof of Proposition \ref{prop:1}}\label{sec:Ap1}

\begin{enumerate}
\item We consider a special channel 
\begin{align*}
    P^{\max}_{\tilde{X}|X,A}(\tilde{x}|x,a) = \frac{1}{|\mathcal{X}|}\ \forall x,\tilde{x}\in\mathcal{X},\forall a.
\end{align*}
We denote the corresponding output as $\hat{Y}_{F_{\max}}$. In this case, $\tilde{X}$ and $X$ are independent. Due to the graphical model in Figure~\ref{fig:GM} and the data processing inequality, $\hat{Y}_F$ and $A$ are also independent which implies that 
\begin{align*}
    \fdiv(P(\hat{Y}_{F_{\max}}|Y,A=0)||P(\hat{Y}_{F_{\max}}|Y,A=1)) = 0.
\end{align*}
We choose $D' = \EE{d(Y,\hat{Y}_{F_{\max}})}$, which is equal to
\begin{align*}
    D' =& \frac{1}{|\mathcal{X}|} \sum_{\tilde{x},y,\Hat{y}_F}W_{Y|X}(\Hat{y}_F|\tilde{x})P_Y(y)d(y,\Hat{y}_F) 
\end{align*}

By the definition, we have $\mathsf{Disc}^{\mathsf{pre|A}}_f(W,D) = 0$ for any $D\geq D'$, and we have $D_{\max}^{\mathsf{pre|A}}\leq D'$. 

With the same argument, we can choose the channel 
\begin{align*}
    P^{\max}_{\Hat{Y}_P|\Hat{Y}_O,A}(\Hat{y}_P|\Hat{y}_O,a) = \frac{1}{|\mathcal{Y}|}, \forall \Hat{y}_P,\Hat{y}_O,a.
\end{align*}
This results to the zero discrimination in the output. $D_{\max}^{\mathsf{post}}$ is upper bounded by the output distortion corresponding to this post-processing channel.

\item
The smallest $D$ such that the constraint in \eqref{eq:opt_five} is feasible can be achieved with a pre-processing channel which results in the smallest possible value for $\EE{d(Y,\Hat{Y}_F)}$. We can make the assignment to the pre-processing channel $P_{\tilde{X}|X,A}$ such that we achieve the smallest possible value for $\EE{d(Y,\Hat{Y}_F)}$. We have
\begin{align*}
\EE{d(Y,\Hat{Y}_F)}=&\sum_{x,\tilde{x},y,\hat{y}_F,a} P_{\tilde{X}|X,A}(\tilde{x}|x,a)W_{Y|X}(\hat{y}_F|\tilde{x})P_{X,Y,A}(x,y,a)d(y,\hat{y}_F)\\
=&\sum_{x,\tilde{x},a}P_{\tilde{X}|X,A}(\tilde{x}|x,a) \sum_{y,\hat{y}_F}W_{Y|X}(\hat{y}_F|\tilde{x})P_{X,Y,A}(x,y,a)d(y,\hat{y}_F)
\end{align*}
In assignment of $P_{\tilde{X}|X,A}$, for every $x$, if we choose $\tilde{x}$, such that $\sum_{y,\hat{y}_F}W_{Y|X}(\hat{y}_F|\tilde{x})P_{X,Y,A}(x,y,a)d(y,\hat{y}_F)$ attain its smallest value, we can claim the resulting distortion from such channel is the smallest feasible distortion. The $P_{\tilde{X}|X,A}$ which is given in Proposition \ref{prop:1} satisfies this property.

The same argument can be done for $D_{\min}^{\mathsf{post}}$.

\item Consider any $D_1, D_2 \geq 0$ and $0\leq\lambda \leq 1$. Let $P^i_{\tilde{X}|X,A}$ result in $\hat{Y}_{F_i}$ and achieve $\mathsf{Disc}^{\mathsf{pre}|A}_f(W,D_i), i=1,2$. In other words,
\[
\mathsf{Disc}^{\mathsf{pre}|A}_f(W,D_i) = \fdiv(P(\hat{Y}_{F_i}|Y,A=0)||P(\hat{Y}_{F_i}|Y,A=1)),
\]
where 
\[
\EE{d(Y;\Hat{Y}_{F_i})} \leq D_i.
\]
Consider the channel $P^\lambda_{\tilde{X}|X,A} = \lambda P^1_{\tilde{X}|X,A} + \Bar{\lambda}P^2_{\tilde{X}|X,A}$, where $\bar{\lambda} = 1-\lambda$. Let the corresponding  output be $\hat{Y}_{F_\lambda}$. Then we have
\begin{align*}
    \EE{d(Y;\Hat{Y}_{F_\lambda})}&= \lambda \EE{d(Y;\Hat{Y}_{F_1})} +\Bar{\lambda}\EE{d(Y;\Hat{Y}_{F_2})}\\    
    &= \lambda D_1 + \Bar{\lambda} D_2 \\
    &= D_\lambda.
\end{align*}
Additionally, we have:
\begin{align*}
    &\lambda \mathsf{Disc}^{\mathsf{pre|A}}_f(W,D_1)+\Bar{\lambda} \mathsf{Disc}^{\mathsf{pre|A}}_f(W,D_2)\\
    =&\lambda \fdiv(P(\hat{Y}_{F_1}|Y,A=0)||P(\hat{Y}_{F_1}|Y,A=1))\\
    & +\Bar{\lambda}\fdiv(P(\hat{Y}_{F_2}|Y,A=0)||P(\hat{Y}_{F_2}|Y,A=1))\\
    \geq &\fdiv(P(\hat{Y}_{F_\lambda}|Y,A=0)||P(\hat{Y}_{F_\lambda}|Y,A=1))\\
    \geq&\mathsf{Disc}^{\mathsf{pre|A}}_f(W,D_\lambda)
\end{align*}

Since $P_{\Hat{Y}_P|Y,A}$ can written as a linear function of the post-processing channel $P_{\Hat{Y}_P|\hat{Y}_O,A}$, we can use the exact same argument to prove the convexity of $\mathsf{Disc}^{\mathsf{post}}_f(W,D)$ with respect to $D$.

\item Since bigger $D$ corresponds to bigger feasible region for the optimization problem in \eqref{eq:opt_one} and \eqref{eq:opt_three}, the function $\mathsf{Disc}^{\mathsf{pre|A}}_f(W,D)$ is a non-increasing function function with respect to $D$ for a fixed $W$. Furthermore, \eqref{eq:opt_one} has a positive minimum for some $D<D_{\max}^{\mathsf{pre|A}}$. In addition, $\mathsf{Disc}^{\mathsf{pre|A}}_f(W,D_{\max}^{\mathsf{pre|A}})=0$. Also, $\mathsf{Disc}^{\mathsf{pre|A}}_f(W,D)$ is convex with respect to $D$. Hence, $\mathsf{Disc}^{\mathsf{pre|A}}_f(W,D)$ is strictly decreasing for $D\leq D_{\max}^{\mathsf{pre|A}}$.

The exact same argument holds for $\mathsf{Disc}^{\mathsf{post}}_f(W,D)$.
\item $P_{\tilde{X}|X,A}(\tilde{x}|x,a) = \delta_{\tilde{x},x}$ results in the output $\Hat{Y}_F$ being exactly the same as $\Hat{Y}_O$, which results in the stated inequality.
\end{enumerate}

\section{Proof of Proposition \ref{prop:substitution}} \label{sec:nonexistence}
\textbf{If part:}
First we show that the value of the objective function and distortion of pre and post processing can be found uniquely by $P_{\Hat{Y}_F|Y,A}$ and $P_{\Hat{Y}_P|Y,A}$, respectively. We have
\[
P_{\Hat{Y}_P,Y} (\Hat{y}_P,y) = \sum_{a}P_{\Hat{Y}_P|Y,A}(\Hat{y}_P|y,a)P_{Y,A}(y,a).
\]
One can calculate $\EE{d(Y,\Hat{Y}_P)}$ uniquely using this joint distribution. In addition, $\EE{D_f(P_{\Hat{Y}_P|Y,A=0}||P_{\Hat{Y}_P|Y,A=1})}$ can be found uniquely using $P_{\Hat{Y}_P|Y,A}$. The exact same argument holds for $P_{\Hat{Y}_F|Y,A}$. 

Given the assumption of the theorem, We need to prove that, if there exist $P_{\Hat{Y}_P|\Hat{Y}_O,A}$ which gives 

\[
P_{\Hat{Y}_P|Y,A}(1|y,a) = \sum_{\Hat{y}_O,x} P_{\Hat{Y}_P|\Hat{Y}_O,A}(1|\Hat{y}_O,a)W_{Y|X}(\Hat{y}_O|x)P_{X|Y,A}(x|y,a),
\]
then there exist $P_{\tilde{X}|X,A}$, such that 
\begin{align*}
P_{\Hat{Y}_F|Y,A}(1|y,a) =& \sum_{x,\tilde{x}} W_{Y|X}(1|\tilde{x})P_{\tilde{X}|X,A}(\tilde{x}|x,a)P_{X|Y,A}(x|y,a) \\
=& P_{\Hat{Y}_P|Y,A}(1|y,a), \forall y,a.
\end{align*}
We have
\[
\sum_{\Hat{y}_O,x} P_{\Hat{Y}_P|\Hat{Y}_O,A}(1|\Hat{y}_O,a)W_{Y|X}(\Hat{y}_O|x)P_{X|Y,A}(x|y,a)=\sum_{x}P_{X|Y,A}(x|y,a) \sum_{\Hat{y}_O}P_{\Hat{Y}_P|\Hat{Y}_O,A}(1|\Hat{y}_O,a)W_{Y|X}(\Hat{y}_O|x),
\]
and
\[
\sum_{x,\tilde{x}} W_{Y|X}(1|\tilde{x})P_{\tilde{X}|X,A}(\tilde{x}|x,a)P_{X|Y,A}(x|y,a) =\sum_{x}P_{X|Y,A}(x|y,a) \sum_{\tilde{x}}W_{Y|X}(1|\tilde{x})P_{\tilde{X}|X,A}(\tilde{x}|x,a).
\]
As a result, if we can find $P_{\tilde{X}|X,A}$ such that
\begin{equation}\label{eq:thmproof}
\sum_{\Hat{y}_O}P_{\Hat{Y}_P|\Hat{Y}_O,A}(1|\Hat{y}_O,a)W_{Y|X}(\Hat{y}_O|x)=\sum_{\tilde{x}}W_{Y|X}(1|\tilde{x})P_{\tilde{X}|X,A}(\tilde{x}|x,a), \forall a,x,    
\end{equation}
we can claim the proof. 

For a given $x$ and $a$, using the assumption of the theorem, first we can choose

\[
P_{\tilde{X}|X,A}^{(0)}(\tilde{x}|x,a) =
\begin{cases}
   1 & \text{for }\tilde{x} = x_0\\    
  0 & \text{o.w.}
\end{cases}
\]
This assignment gives $\sum_{\tilde{x}}W_{Y|X}(1|\tilde{x})P_{\tilde{X}|X,A}^{(0)}(\tilde{x}|x,a) = 0$. Second, we can choose
\[
P_{\tilde{X}|X,A}^{(1)}(\tilde{x}|x,a) =
\begin{cases}
   1 & \text{for }\tilde{x} = x_1\\    
  0 & \text{o.w.,}
\end{cases}
\]
and we get $\sum_{\tilde{x}}W_{Y|X}(1|\tilde{x})P_{\tilde{X}|X,A}^{(1)}(\tilde{x}|x,a) = 1$. Since the left hand side of Equation \eqref{eq:thmproof} is always between zero and one, we can assign $P_{\tilde{X}|X,A}$ as a convex combination of $P_{\tilde{X}|X,A}^{(0)}$ and $P_{\tilde{X}|X,A}^{(1)}$, such that the equality in Equation \eqref{eq:thmproof} holds. This assignment can be done for all $x$ and $a$, which proves the sufficiency part.

\textbf{Only if part:}
Assume a trivial post-processor $P_{\Hat{Y}_P|\Hat{Y}_O,A}(1|\Hat{y}_O,a) = 1, \forall \Hat{y}_O,a$. This post-processor generates $1$ in the output, regardless of the output of $W$ and the protected attribute. It is easy to observe that, in order to substitute this post-processor with a pre-processor, we have to have $x_1$ such that $W_{Y|X}(1|x_1) = 1$, so that the pre-processor can map all $x$ to $x_1$ and the resulting $\Hat{Y}_F$ will be constant $1$. If such $x_1$ does not exist, with every pre-processor we always get $\Hat{Y}_F=0$ with a nonzero probability. 

The same argument holds for the case of having $P_{\Hat{Y}_P|\Hat{Y}_O,A}(0|\Hat{y}_O,a) = 1, \forall \Hat{y}_O,a$, as the post-processor, which requires us to have $x_0$, such that $W_{Y|X}(1|x_0) = 0$, or $W_{Y|X}(0|x_0) = 1$.

\section{Proof of Proposition \ref{prop:10}}\label{sec:AP20}
\begin{figure}
    \centering
    \includegraphics[scale=0.4]{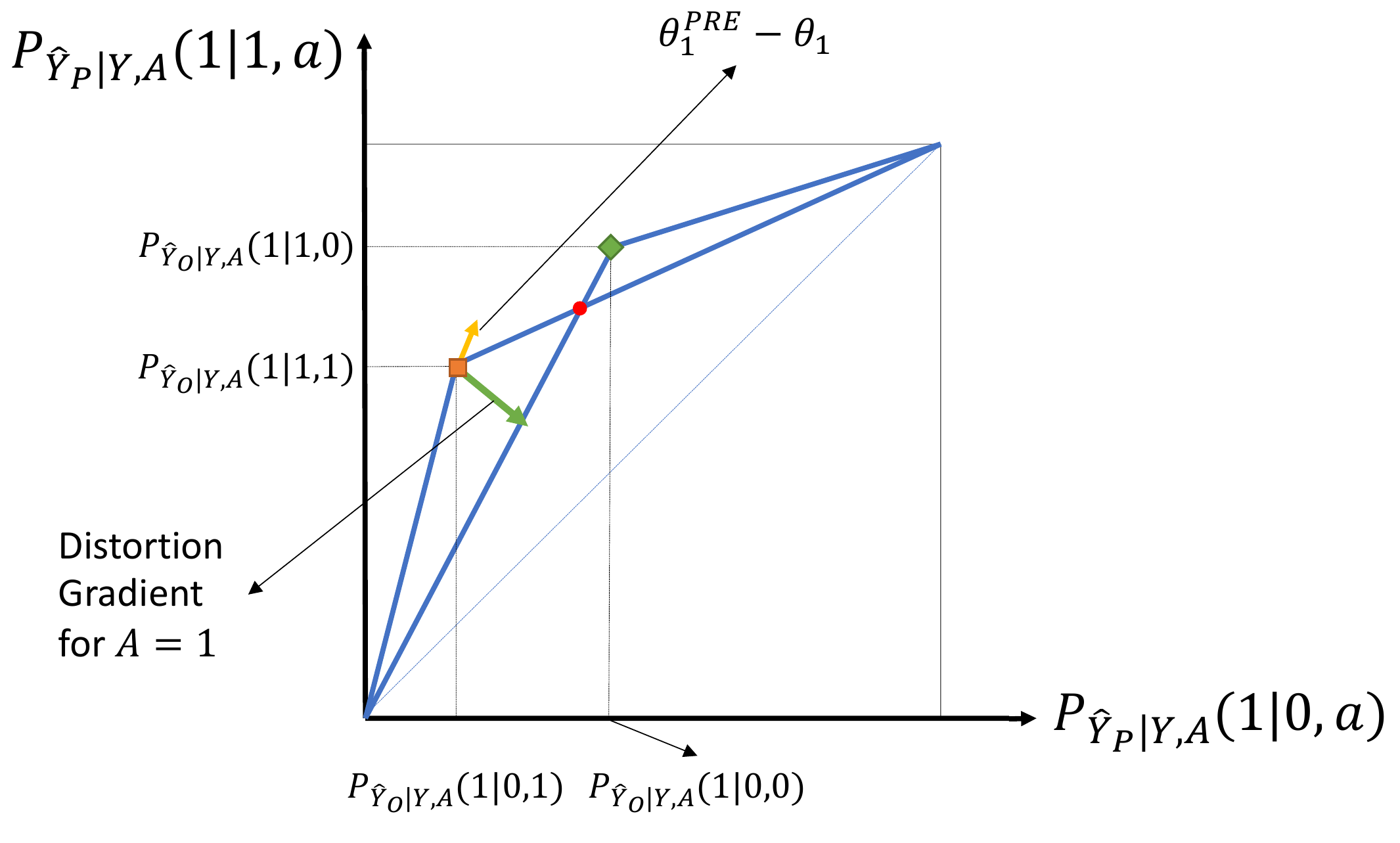}
    \caption{Detection-False alarm graph of the prediction outcome.}
    \label{fig:region}
\end{figure}

For the purpose of the proof, we define (look at Figure \ref{fig:region})
\begin{align*}
\theta_a &= (P_{\Hat{Y}_O|Y,A}(1|0,a),P_{\Hat{Y}_O|Y,A}(1|1,a)),\\
\theta_a^{\mathsf{post}} &= (P_{\Hat{Y}_P|Y,A}(1|0,a),P_{\Hat{Y}_P|Y,A}(1|1,a)),\\
\theta_a^{\mathsf{pre}} &= (P_{\Hat{Y}_F|Y,A}(1|0,a),P_{\Hat{Y}_F|Y,A}(1|1,a)).
\end{align*}

Assume $W$ is a proper classifier. First we will prove that the minimum possible distortion for the post-processing method can be achieved by the trivial post-processing channel $P_{\hat{Y}_P|\hat{Y}_O,A}(\hat{y}_P|\hat{y}_O,a) = \mathds{1}_{\hat{y}_P=\hat{y}_O}$. 

The distortion in Equation \eqref{eq:opt_five} can be written as 
\begin{align*}
\EE{d(Y,\Hat{Y}_P)}
=&P_{\Hat{Y}_P|Y,A}(0|1,0)P_{Y,A}(1,0)\\
&+P_{\Hat{Y}_P|Y,A}(1|0,0)P_{Y,A}(0,0)\\
&+P_{\Hat{Y}_P|Y,A}(1|0,1)P_{Y,A}(0,1)\\
&+P_{\Hat{Y}_P|Y,A}(0|1,1)P_{Y,A}(1,1)\\
=&-P_{\Hat{Y}_P|Y,A}(1|1,0)P_{Y,A}(1,0)\\
&+P_{\Hat{Y}_P|Y,A}(1|0,0)P_{Y,A}(0,0)\\
&+P_{\Hat{Y}_P|Y,A}(1|0,1)P_{Y,A}(0,1)\\
&-P_{\Hat{Y}_P|Y,A}(1|1,1)P_{Y,A}(1,1)\\
&+P_{Y,A}(1,0) + P_{Y,A}(1,1)\\
=& \sum_a \left[P_{\Hat{Y}_P|Y,A}(1|0,a)P_{Y,A}(0,a)- P_{\Hat{Y}_P|Y,A}(1|1,a)P_{Y,A}(1,a)\right]\\
&+P_{Y,A}(1,0) + P_{Y,A}(1,1)\\
=&\sum_a \theta_a^{\mathsf{post}}.(P_{Y,A}(0,a),-P_{Y,A}(1,a))\\
&+P_{Y,A}(1,0) + P_{Y,A}(1,1)\\
\end{align*}
which is a function of $\theta_a^{\mathsf{post}}, a\in\{0,1\}$. We have
\begin{align}\label{eq:nabla}
\nabla_{\theta_a^{\mathsf{post}}}\EE{d(Y,\Hat{Y}_P)} = \left(P_{Y,A}(0,a),-P_{Y,A}(1,a)\right).
\end{align}
We know that all the achievable points $\theta_a^{\mathsf{post}}$ in the Detection-False alarm graph is inside the triangle defined by three points $(0,0)$, $(1,1)$, $\theta_a$. (Look at figure \ref{fig:region}). The slope of the line between $(1,1)$ and $\theta_a$ is equal to $\frac{P_{\Hat{Y}_O|Y,A}(0|1,a)}{P_{\Hat{Y}_O|Y,A}(0|0,a)}$, and the slope of the line between $\theta_a$ and $(0,0)$ is equal to $\frac{P_{\Hat{Y}_O|Y,A}(1|1,a)}{P_{\Hat{Y}_O|Y,A}(1|0,a)}$. Furthermore, the slope of the prependicular line to the $\nabla_{\theta_a^{\mathsf{post}}}\EE{d(Y,\Hat{Y}_P)}$ is equal to $\frac{P_{Y,A}(0,a)}{P_{Y,A}(1,a)}$. Hence, if we have 
\[
\frac{P_{\Hat{Y}_O|Y,A}(1|1,a)}{P_{\Hat{Y}_O|Y,A}(1|0,a)}>\frac{P_{Y,A}(0,a)}{P_{Y,A}(1,a)}>\frac{P_{\Hat{Y}_O|Y,A}(0|1,a)}{P_{\Hat{Y}_O|Y,A}(0|0,a)},
\]
any point in the triangle $(0,0)$, $(1,1)$, $\theta_a$, will have higher distortion than the original point $\theta_a$. These conditions can be simplified as 
\[
P_{\Hat{Y}_O,Y,A}(0,0,a)>P_{\Hat{Y}_O,Y,A}(0,1,a),
\]
and
\[
P_{\Hat{Y}_O,Y,A}(1,1,a)>P_{\Hat{Y}_O,Y,A}(1,0,a),
\]
which are the conditions of a proper classifier.

In addition, having $P_{\tilde{X}|X,A}(\tilde{x}|x,a) = \mathds{1}_{\tilde{x}=x}$, results in $\theta_a^{\mathsf{pre}} = \theta_a$, and for an arbitrary $P_{\tilde{X}|X,A}(\tilde{x}|x,a)$, we have   
\[
P_{\Hat{Y}_F|Y,A}(1|y,a) = \sum_{x,\tilde{x}} W_{Y|X}(1|\tilde{x})P_{\tilde{X}|X,A}(\tilde{x}|x,a)P_{X|Y,A}(x|y,a). 
\]
Consider the following pre-processor
\begin{equation}\label{eq:app9}
P_{\tilde{X}|X,A}(\tilde{x}|x,a) =
\begin{cases}
  1 & \text{if}~\tilde{x} = x, A=0\\
  1 & \text{if}~\tilde{x} = x, x\neq x_a, A=1\\    
  (1-\alpha) & \text{if}~ \tilde{x} = x = x_i,A=1\\
  \alpha & \text{if}~ \tilde{x} = x_{max},  x = x_i,A=1\\
  0 & \text{o.w.}
\end{cases}
\end{equation}

Such an assignment results in 
\[
P_{\Hat{Y}_F|Y,A}(1|y,0) =P_{\Hat{Y}_O|Y,A}(1|y,0), 
\]
and
\begin{equation}\label{eq:app9.5}
P_{\Hat{Y}_F|Y,A}(1|y,1) =P_{\Hat{Y}_O|Y,A}(1|y,1) +\alpha P_{X|Y,A}(x_a|y,1)[W_{Y|X}(1|x_{\max})-W_{Y|X}(1|x_i)],
\end{equation}

Since $[W_{Y|X}(1|x_{\max})-W_{Y|X}(1|x_i)] > 0$, this pre-processor results in $\theta_0^{\mathsf{pre}} = \theta_0$, which has the same false alarm and detection as the original classifier, and $\theta_1^{\mathsf{pre}}$, which has bigger false alarm and detection than the original classifier. Since we made the natural assumption that the majority group has higher detection and higher false alarm, this assignment for the pre-processor results in $\theta_0^{\mathsf{pre}}$ and $\theta_1^{\mathsf{pre}}$ which compared to $\theta_0$ and $\theta_1$ are closer in terms of false alarm and detection, and results in a lower discrimination. 

From Equation \eqref{eq:app9.5} we have
\begin{equation}\label{eq:app10}
\frac{P_{X|Y,A}(x_i|0,1)}{P_{X|Y,A}(x_i|1,1)} = \frac{P_{\Hat{Y}_F|Y,A}(1|0,1) -P_{\Hat{Y}_O|Y,A}(1|0,1) }{P_{\Hat{Y}_F|Y,A}(1|1,1) -P_{\Hat{Y}_O|Y,A}(1|1,1)}.    
\end{equation}

Assuming $P_{Y|X,A}(0|x_i,1)<P_{Y|X,A}(1|x_i,1)$, we have
\begin{align}
P_{Y,X,A}(0,x_i,1)&<P_{Y,X,A}(1,x_i,1)\nonumber\\
\implies \frac{P_{X|Y,A}(x_a|0,1)}{P_{X|Y,A}(x_a|1,1)} &< \frac{P_{Y,A}(1,1)}{P_{Y,A}(0,1)}.\label{eq:app11}
\end{align}
Combining \eqref{eq:app10} and \eqref{eq:app11}, we get
\begin{align}\label{eq:slope_slope}
\frac{P_{\Hat{Y}_F|Y,A}(1|0,1) -P_{\Hat{Y}_O|Y,A}(1|0,1) }{P_{\Hat{Y}_F|Y,A}(1|1,1) -P_{\Hat{Y}_O|Y,A}(1|1,1)} < \frac{P_{Y,A}(1,1)}{P_{Y,A}(0,1)}.
\end{align}
By following the same argument as in post-processor in Equation \eqref{eq:nabla}, we get that $\frac{P_{Y,A}(0,1)}{P_{Y,A}(1,1)}$ is the slope of the perpendicular of the gradient of $\EE{d(Y,\hat{Y}_F)}$ with respect to $\theta_1^{\mathsf{pre}}$. Furthermore, $\frac{P_{\Hat{Y}_F|Y,A}(1|1,1) -P_{\Hat{Y}_O|Y,A}(1|1,1)}{P_{\Hat{Y}_F|Y,A}(1|0,1) -P_{\Hat{Y}_O|Y,A}(1|0,1)}$ is the slope of the line $\theta_1^{\mathsf{pre}}-\theta_1$ (Look at figure \ref{fig:region}). Inequality \eqref{eq:slope_slope} implies that $\theta_1^{\mathsf{pre}}-\theta_1$ points to the opposite direction of the gradient of the distortion, and hence the distribution assignment in \eqref{eq:app9} results in $\hat{Y}_F$ which has a lower distortion, say $D'$, compared to $D_{\min}^{\mathsf{post}}$. 

If there exist $x_b$, such that $P_{Y|X,A} (1|x_b,0) < P_{Y|X,A} (0|x_b,0)$, one can make the same argument and choose a pre-processor that has a lower distortion compared to the lowest achievable distortion via post-processing, and at the same time has smaller discrimination compared to the discrimination of $\Hat{Y}_O$. 

As a result, we can find a pre-processor that has distortion $D'<D_{\min}^{\mathsf{post}}$, and $\mathsf{Disc}_f^{\mathsf{pre|A}}(W,D') < \mathsf{Disc}_f^{\mathsf{post}}(W,D_{\min})$. Since $\mathsf{Disc}_f^{\mathsf{pre|A}}(W,D)$ is a decreasing function of $D$, and $D'<D_{\min}^{\mathsf{post}}$, we have \[
\mathsf{Disc}_f^{\mathsf{pre|A}}(W,D_{\min}^{\mathsf{pre|A}}) < \mathsf{Disc}_f^{\mathsf{post}}(W,D_{\min}^{\mathsf{post}}).
\]

\section{}\label{sec:AP2}
We prove that the more restricted distortion constraint $\EE{d(Y;\Hat{Y}_F)|X=x} \leq D~~\forall x$ results in a linear constraint with respect to $P_{\tilde{X}|X}$. Using Figure~\ref{fig:GM}, we have
\begin{align*}
    \EE{d(Y,\hat{Y}_F)|X=x}
    =&\sum_{y,\hat{y}_F}P_{Y,\Hat{Y}_F|X}(y,\hat{y}_F|x)d(y,\hat{y}_F)\\
    = &\sum_{\tilde{x},y,\hat{y}_F} P_{\tilde{X}|X}(\tilde{x}|x)W_{Y|X}(\hat{y}_F|\tilde{x})P_{Y|X}(y|x)d(y,\hat{y}_F), 
\end{align*}
As a result, the conditional distortion constraint is linear with respect to $P_{\tilde{X}|X}$. All the other properties simply follows from the linearity.

\section{Demographic Parity as the Fairness Criterion}\label{sec:AP3}
In The demographic parity fairness criterion, the objective function to minimize is
\begin{align}\label{eq:EOObj}
    \EE{\fdiv\big(P_{\hat{Y}_F|A=0}\|P_{\hat{Y}_F|A=1}\big)}.
\end{align}

According to the graphical model in figure \ref{fig:GM} We have

\begin{align*}
    P_{\hat{Y}_F|A} (\Hat{y}_F|a)&= \sum_{x,\tilde{x}}P_{\hat{Y}_F,X,\tilde{X}|A}(\hat{y}_F,x,\tilde{x}|a)\\
    &=\sum_{x,\tilde{x}}W_{Y|\tilde{X}}(\hat{y}_F|\tilde{x})P_{\tilde{X}|X}(\tilde{x}|x)P_{X|A}(x|a), ~~\forall \Hat{y}_F,a,
\end{align*}
which is a linear function of $P_{\tilde{X}|X}$. In Appendix \ref{sec:AP0} we have shown that this results to the convexity of $\fdiv\big(P(\hat{Y}_F|A=0)\|P(\Hat{Y}_F|A=1)\big)$ with respect to $P(\tilde{X}|X)$. Having the convexity, the rest of the properties can be derived similarly.

\section{Proof of lemma \ref{lem:poly}} \label{sec:AP5}
Using sensitivity analysis in linear programming \cite{bertsimas1997introduction}, we can write \eqref{eq:LP} as
\begin{equation}\label{eq:LP-simple}
\begin{split}
  \mathsf{Disc}^{\mathsf{pre|A}}_{TV}(\W,D)=\min_{z}~~ &c^Tz,\\
  \text{s.t}: A&z = b,\\
  a^T&z= D,\\
  z\geq& 0,
\end{split}
\end{equation}
where $z$ is vector obtained by concatenating $t_1,\cdots,t_{|\mathcal{Y}|}$, $P_{\Tilde{X}|X,A}(\Tilde{x}|x,a),\forall \tilde{x},x,a$, and $P_{\Hat{Y}_F|Y,A}(\Hat{y}_F|y,a),\forall\Hat{y},y,a$, and necessary slack variables. $c$ contains required coefficient to make the objective function the same as the one in \eqref{eq:LP}. $a^Tz= D$ corresponds to distortion constraint, which have been altered to equality by a slack variable, and $Az= b$ represents all the equities and all the inequalities, except for the aforementioned distortion constraint. By changing $D$ to $D+\delta D$, as long as the basis of the linear solution remains the same, the objective function changes linearly with respect to $\delta D$  \cite[p.~208]{bertsimas1997introduction}. Combining this result with Proposition \ref{prop:1}, we conclude that the function $D\mapsto \mathsf{Disc}^{\mathsf{pre|A}}_f(D,\W)$ is a piecewise linear decreasing convex function. 
The exact same argument holds for $D\mapsto \mathsf{Disc}^{\mathsf{post}}_f(D,\W)$.

\section{The linear program for designing the post-processor}\label{sec:AP40}
Given a distortion upper bound $D$ and the joint distribution $P_{\Hat{Y}_O,Y,A}$, the following linear program can be used to find a post-processing channel $P_{\Hat{Y}_P|\Hat{Y}_O,A}$:
\begin{align*}
  \mathsf{Disc}^{\mathsf{post}}_{TV}(W,D)&=
  \min_{P_{\Hat{Y}_P|\Hat{Y}_O,A},P_{\Hat{Y}_P|Y,A},t}~~ \sum_{y=1}^{|\mathcal{Y}|}P_Y(y)\sum_{\hat{y}_P=1}^{|\mathcal{Y}|} t_{y,\hat{y}_P},\\
  &\text{s.t.}  ~~P_{\Hat{Y}_P|Y,A}(\hat{y}_P|y,0)-P_{\Hat{Y}_P|Y,A}(\hat{y}_P|y,1)  - t_{y,\hat{y}_P}\leq 0, ~~\forall y,\Hat{y}_P,\\
  &P_{\Hat{Y}_P|Y,A}(\hat{y}_P|y,1)-P_{\Hat{Y}_P|Y,A}(\hat{y}_P|y,0)- t_{y,\hat{y}_P}\leq 0, ~~\forall y,\Hat{y}_P,\\
  &\sum_{\Hat{y}_P,\Hat{y}_O,y,a} \big[P_{\Hat{Y}_P|\Hat{Y}_O,A}(\Hat{y}_P|\Hat{y}_O,a) P_{\Hat{Y}_O|Y,A}(\Hat{y}_O|y,a) P_{Y,A}(y,a)d(y,\Hat{y}_P)\big]\leq D\\  
  & P_{\Hat{Y}_P|Y,A}(\Hat{y}_P|y,a)-\sum_{\Hat{y}_O} P_{\Hat{Y}_P|\Hat{Y}_O,A}(\Hat{y}_P|\Hat{y}_O,a)P_{\Hat{Y}_O|Y,A}(\Hat{y}_O|y,a)\! =\! 0, \forall \Hat{y}_P,y,a,\\
  &\sum_{\hat{y}_P} P_{\Hat{Y}_P|Y,A}(\hat{y}_P|y,a) = 1,~~ \forall y, a,  \\
  &\sum_{\Hat{y}_P} P_{\Hat{Y}_P|\Hat{Y}_O,A}(\Hat{y}_P|\Hat{y}_O,a) = 1,~~ ~~ \forall \Hat{y}_O,a,\\
  &P_{\Hat{Y}_P|Y,A}(\hat{y}_P|y,a) \geq 0,~~ \forall \hat{y}_P,y a,\\
  &P_{\Hat{Y}_P|\Hat{Y}_O,A}(\Hat{y}_P|\Hat{y}_O,a)  \geq 0, ~~ \forall \Hat{y}_P,\Hat{y}_O,a
\end{align*}

\end{document}